\documentclass{article}

\usepackage{microtype}
\usepackage{graphicx}
\usepackage{subcaption}
\usepackage{booktabs}
\usepackage{float}
\usepackage{enumitem}
\usepackage{multirow}
\usepackage{pifont}

\graphicspath{{Figures/}}

\usepackage{hyperref}
\usepackage{xurl}

\usepackage[preprint]{icml2026}         

\usepackage{amsmath}
\usepackage{amssymb}
\usepackage{mathtools}
\usepackage{amsthm}

\usepackage[capitalize,noabbrev]{cleveref}

\theoremstyle{plain}
\newtheorem{theorem}{Theorem}[section]

\theoremstyle{definition}

\theoremstyle{remark}
\newtheorem{remark}[theorem]{Remark}

\usepackage[textsize=tiny]{todonotes}

\newcommand{\READ}{\textsc{Read}}
\newcommand{\rupee}{\textrm{Rs.}}

\icmltitlerunning{Beyond Top-K: Interpretable Agentic Operations for Document Retrieval}

\begin{document}
\AddToShipoutPictureFG*{
    \AtPageUpperLeft{
        \hspace*{0.44\paperwidth}
        \raisebox{-2.0cm}{
            \includegraphics[width=3.2cm]{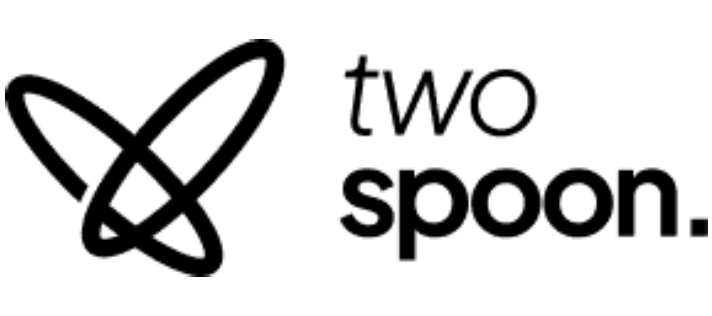}
        }
    }
}

\newsavebox{\teaserbox}
\sbox{\teaserbox}{\includegraphics[width=0.70\textwidth]{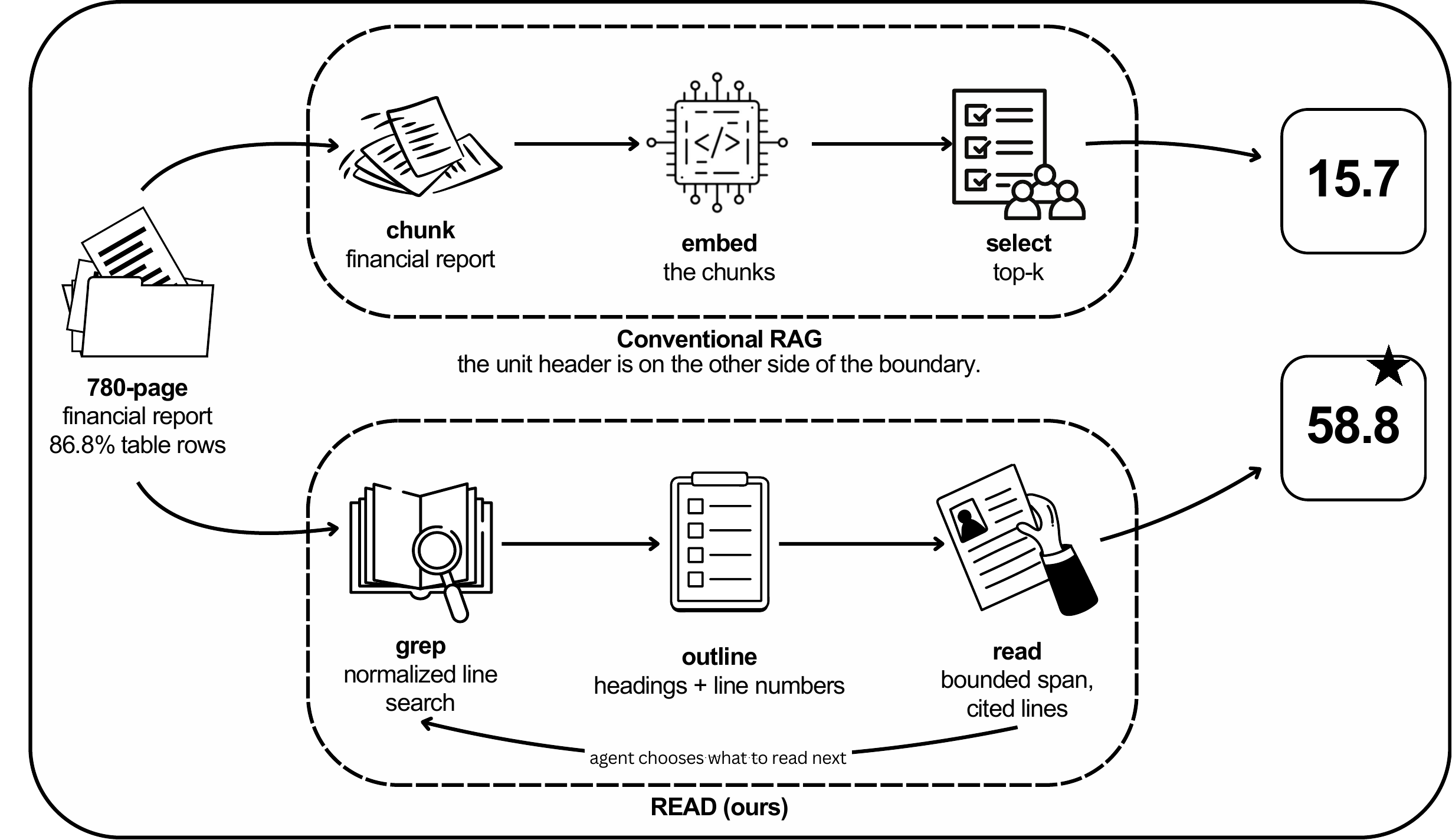}}

\twocolumn[
  \icmltitle{Beyond Top-K: Replacing Black-Box Retrieval with \\
  Interpretable Agentic Operations}

  \begin{icmlauthorlist}
    \icmlauthor{Sagar Tamang}{aff1,aff2}
    \icmlauthor{Ayush Vyas}{aff2}
    \icmlauthor{Tabarakul Hazarika}{aff2}
  \end{icmlauthorlist}

  \icmlaffiliation{aff1}{Indian Institute of Technology Patna, Bihta, India}
  \icmlaffiliation{aff2}{TwoSpoon, India}

  \icmlcorrespondingauthor{Sagar Tamang}{sagar\_pa2508mch184@iitp.ac.in}

  \icmlkeywords{Retrieval-Augmented Generation, Agentic Search, Tool Use,
  Interpretability, Document Understanding}

  \vskip 0.10in
  \begin{center}
    \small Code, benchmark, and all result trajectories:
    \url{https://github.com/twospoon/READ}
  \end{center}

  \vskip 0.12in
  \begin{center}
    \usebox{\teaserbox}
    \captionof{figure}{Two retrieval interfaces over the same document. The
    conventional pipeline commits to a partition \emph{before} the query is
    known, so a figure can arrive without the header that states its unit and
    fiscal year --- on this document that separation survives every chunk size
    we tried (\Cref{sec:corpus}). \READ{} leaves the document intact and lets
    the agent choose what to read next, so each step is a citable line range
    rather than a similarity score. Figures at right are accuracy, in percent
    correct, over the same 51 verified questions: 15.7\% for dense top-$k$
    against 58.8\% for \READ{} (\Cref{sec:results}). Tuning the dense
    baseline's chunk size and retrieval depth raises it to 35.3\% at best,
    which \READ{} still leads by 23.5 points (\Cref{app:sweep}).}
    \label{fig:teaser}
  \end{center}

  \vskip 0.12in
]

\printAffiliationsAndNotice{}

\begin{abstract}
Retrieval-augmented generation over long documents is dominated by one design:
chunk the text, embed the chunks, and surface the top-$k$ nearest neighbours of
the query. We argue that for an important class of documents --- financial
statements, audit reports, regulatory returns --- this design is structurally
unsound, and we make the argument measurable. On a 780-page government financial
report, 86.8\% of content lines are table rows, thousands of near-identical
figures compete in one embedding space, and a figure inherits its unit from a
header a median of 13 lines above it --- so a chunk boundary routinely separates
a number from whether it is in lakh or crore, an error of two orders of
magnitude. A table-aware chunker built as a steelman fixes the unit problem but
leaves 27--30\% of numeric chunks with no fiscal-year header at \emph{every}
chunk size we tried. We propose \READ{} (Reliable Embedding-free Agentic
Document-search), in which an agent reads the raw document through three
deterministic operations --- normalized lexical search, structural navigation,
and bounded span reads --- exposed over the Model Context Protocol, so a
trajectory is a replayable audit trail, not an opaque similarity score. On 51
verified questions \READ{} answers 58.8\% against dense retrieval's 15.7\%
($p_{\text{Holm}}=2\times10^{-5}$) --- or 35.3\% tuned, which \READ{} still
leads by 23.5 points ($p_{\text{Holm}}=0.017$). An agent given the same loop but
a top-$k$ tool reaches only 27.5\%, locating the gain in the interface rather
than in iteration. We also report what the evidence does not support: BM25 is
statistically indistinguishable from \READ, so our result separates
embedding-based from embedding-free retrieval, not agentic from lexical search.
\end{abstract}

\section{Introduction}
\label{sec:intro}

Retrieval-augmented generation (RAG) is the default recipe for grounding
language models in external knowledge \cite{lewis2020rag,gao2023ragsurvey}. A
corpus is split into chunks, each is mapped to a dense vector
\cite{karpukhin2020dpr}, and at query time the top-$k$ nearest neighbours are
concatenated into the prompt. A decade of work has refined every stage ---
chunking policies, embedding models, sparse--dense fusion, reranking
\cite{wang2024ragbestpractices,cormack2009rrf,khattab2020colbert} --- but the
contract has not changed: the model sees a small, similarity-ranked slice of the
corpus, selected by a mechanism it cannot inspect, steer, or verify.

This paper is about a class of documents for which that contract fails, and
about why the failure is structural rather than a matter of tuning. The class is
large: financial statements, audit reports, statutory filings, and regulatory
returns, which share three properties --- they are overwhelmingly tabular, they
contain thousands of near-duplicate numbers, and they carry meaning in layout,
where a figure's units and fiscal year are declared in a header some distance
above it.

Consider the question that motivated this work, asked of the \emph{Gujarat
Finance Accounts 2024--25, Volume I}: \emph{what is the difference between the
Fiscal Deficit and the Revenue Surplus for 2024--25?} The two operands sit 47
lines apart in the same statement, and their difference appears nowhere in the
document --- it must be computed. In a pilot run (\Cref{sec:pilot}), a tuned
dense retriever failed to place both operands in the same top-$k$ context on
any trial, and the generator correctly reported that it could not answer. This
is not a hallucination; it is retrieval declining to deliver, and it is the
best case. The worse case is a retriever that surfaces one plausible figure and
a fluent generator that uses it.

We make three moves. First, we \emph{measure} the setting rather than assert it:
\Cref{sec:corpus} quantifies tabularity, numeric near-duplication, and the
distance between a figure and its governing unit header, and then measures how
much of that context chunking destroys. Second, we build the strongest chunker
we can --- one that carries unit and fiscal-year headers across every boundary
--- so that what remains is a property of the paradigm rather than of a lazy
implementation. Third, we describe \READ{} --- \emph{Reliable Embedding-free Agentic
Document-search} --- an interface in which a tool-calling agent reads the
document directly, and we compare it against dense, sparse, and hybrid
baselines under a paired protocol.

Our contributions:

\begin{itemize}[leftmargin=*,itemsep=2pt]
  \item \textbf{A measured characterisation of exactness-critical documents.}
        We quantify the properties that defeat chunk-and-embed retrieval on a
        real 780-page financial report, including a unit-inheritance analysis we
        have not seen reported elsewhere (\Cref{sec:corpus}).
  \item \textbf{A steelmanned baseline, and the residue it cannot fix.}
        Table-aware chunking cuts unitless chunks from 18.0\% to 0.3\%, but
        28.9\% of numeric chunks still carry no fiscal-year header
        (\Cref{tab:chunking}). We report the strong configuration throughout.
  \item \textbf{\READ, an embedding-free retrieval interface}, released as an
        MCP server, together with the finding that \emph{normalized} lexical
        matching is what makes it viable on converted PDFs --- and a measurement
        of exactly which conversion artifacts make it so (\Cref{sec:normalization}).
  \item \textbf{A conversion-fidelity ceiling.} We quantify the damage the
        PDF converter inflicts (\Cref{tab:conversion}) and show it is a bound
        shared by every retrieval method, requiring its own question category
        rather than being attributed to a retriever (\Cref{sec:conversion}).
  \item \textbf{An evaluation protocol with three independent grading signals},
        including a mechanical groundedness check that a fluent-but-unsupported
        answer cannot pass (\Cref{sec:metrics}).
\end{itemize}

\section{Related Work}
\label{sec:related}

\subsection{RAG and Its Moving Parts}

RAG couples a retriever with a generator so outputs are conditioned on retrieved
evidence \cite{lewis2020rag}. Dense passage retrieval established learned dual
encoders as the dominant retriever \cite{karpukhin2020dpr}, refined by
late-interaction models \cite{khattab2020colbert}, learned sparse expansion
\cite{formal2021splade}, and unsupervised contrastive encoders
\cite{izacard2022contriever}. The BEIR benchmark showed that BM25
\cite{robertson1995bm25,thakur2021beir} remains a strong zero-shot baseline
against far heavier neural retrievers --- an early signal that lexical matching
is harder to beat than assumed.

In deployment the retriever is one of many coupled components.
\citet{wang2024ragbestpractices} catalogue the design space and find end-to-end
quality highly sensitive to each choice. Two fragilities matter here. Position
effects: models under-use evidence in the middle of long contexts
\cite{liu2024lost}, so a retriever that surfaces the right chunk may still not
deliver it usefully. Behavioural drift: the services powering embedding and
generation change opaquely over time \cite{chen2023chatgptdrift}, so an index
built against one model version silently degrades against another. Adaptive
schemes such as Self-RAG teach the model when to retrieve and to critique what
it retrieved \cite{asai2023selfrag}, but keep the index and its maintenance
burden.

\subsection{Long and Structured Documents}

Long-document QA exposes the weakest link of chunk-based RAG: the chunker.
Financial statements encode meaning in layout --- row/column adjacency,
multi-page tables, header inheritance --- which fixed-size chunking destroys.
Work on spreadsheets illustrates the escalating machinery required to
compensate: SpreadsheetLLM compresses sheets to fit context windows
\cite{dong2024spreadsheetllm}, while FRTR builds multi-granular row, column, and
block embeddings fused by reciprocal rank fusion to reach usable accuracy on
enterprise workbooks \cite{gulati2026frtr}. These systems show that
structure-aware retrieval \emph{can} be built, at the price of ever more
elaborate corpus-specific indexing. We take the opposite route: leave the
document intact and give the agent operations expressive enough to exploit its
structure directly.

An alternative response is to dispense with retrieval and rely on long context.
Our document is 1.68\,MB and fits a 1M-token window, so this is feasible here;
but it remains costly, exhibits positional degradation \cite{liu2024lost}, and
does not scale to corpora that keep growing. We include it as a baseline rather
than dismissing it.

\subsection{Agentic Retrieval and Tool Use}

A separate line makes retrieval \emph{iterative}. Multi-hop systems learn to
retrieve reasoning paths rather than isolated passages
\cite{asai2020path,trivedi2023ircot}, and ReAct-style agents interleave
reasoning with search \cite{yao2023react,schick2023toolformer}. The competence
of models at operating software \cite{chen2021codex} underwrites a stronger
version: SWE-agent showed a bash-equipped agent navigates million-line
repositories \cite{yang2024sweagent}, and Agentless showed a fixed
grep-and-read pipeline rivals full agent frameworks \cite{xia2024agentless}.
Memory-augmented conversational agents such as Chronos likewise replace opaque
recall with structured, searchable stores \cite{lumer2026chronos}. The Model
Context Protocol \cite{anthropic2024mcp} standardises how such tools are
exposed. \READ{} is an MCP server whose tools target a single long document
rather than a repository.

\subsection{Direct Corpus Access versus Dense Retrieval}

Closest to this work are two recent comparisons. \citet{sen2026grep} evaluate
grep against vector search across custom and provider-native agent harnesses on
conversational-memory QA, finding inline grep beats vector retrieval for every
harness--model pair tested, and that the harness shifts accuracy as much as the
retriever does. \citet{dci2026interface} formalise \emph{direct corpus
interaction} --- replacing the retrieval API with terminal primitives over the
raw corpus --- improving accuracy on BrowseComp-Plus while reducing cost, and
introduce \emph{retrieval interface resolution} as the explanatory lens.
Practitioner accounts report the same pattern \cite{yu2025grepvsgraph}.

We differ in four ways.
\textbf{Domain.} Prior comparisons target multi-document corpora --- web
snapshots, conversation logs, repositories --- where the task is finding the
right file. We target a single, very long, densely tabular document, where the
task is navigating structure \emph{within} one file.
\textbf{Task.} Our benchmark is exactness-critical: answers are specific
figures whose correctness is binary, and several must be computed rather than
retrieved.
\textbf{Corpus condition.} Prior work greps clean text. We grep the output of a
PDF converter, and show in \Cref{sec:normalization} that naive lexical matching
is materially degraded by conversion artifacts --- a failure mode that does not
arise on clean corpora and that, to our knowledge, has not been reported.
\textbf{Framing.} Beyond accuracy and cost we foreground interpretability:
because every operation is deterministic, the trajectory is an audit log, which
is a requirement in the financial-reporting settings we study.

\section{Two Retrieval Interfaces}
\label{sec:interfaces}

We frame the comparison as one between two \emph{interfaces} to the same
document, consumed by the same backbone. Let $D$ be a document rendered to text
as an ordered sequence of lines $D=(\ell_1,\dots,\ell_L)$ with a structural map
$S(D)$ and a unit map $u:\{1..L\}\to\mathcal{U}$ giving the unit in force at
each line.

\paragraph{Interface A: embedding-mediated top-$k$.}
An offline pipeline partitions $D$ into chunks $\{c_j\}_{j=1}^{M}$, embeds each
as $\mathbf{v}_j=E(c_j)\in\mathbb{R}^d$, and returns
\begin{equation}
R_k(q)=\operatorname*{arg\,top\text{-}k}_{j\in[M]}
\frac{E(q)\cdot\mathbf{v}_j}{\lVert E(q)\rVert\,\lVert\mathbf{v}_j\rVert}.
\label{eq:topk}
\end{equation}
Three properties follow. \emph{Lossy projection:} all evidence available to
selection is compressed into $d$ floats, so distinctions below the resolution of
$E$ are unrecoverable at selection time. \emph{Fixed budget:} exactly $k$ chunks
are surfaced whether the question needs one cell or forty pages. \emph{Opacity:}
the only account of why $c_j$ was chosen is a scalar.

A fourth property is specific to structured documents and is the one we
quantify. Chunking is a partition of $D$, but $u$ is defined by inheritance from
preceding lines. A chunk $c_j$ spanning lines $[a,b]$ is \emph{unit-preserving}
only if some line in $[a,b]$ declares $u(b)$. Nothing in \Cref{eq:topk} enforces
this, and \Cref{sec:corpus} shows how often it fails.

\paragraph{Interface B: interpretable agentic operations.}
\READ{} exposes $D$ through deterministic operations:
\begin{itemize}[leftmargin=*,itemsep=1pt]
  \item $\textsc{search}(p)\rightarrow\{(i,\ell_i,\mathrm{loc}(i))\}$: all lines
        matching pattern $p$ under a normalization $\nu$
        (\Cref{sec:normalization}), with line number and structural location;
  \item $\textsc{outline}()\rightarrow S(D)$: statements and sections with their
        line and page extents;
  \item $\textsc{read}(a,b)\rightarrow(\ell_a,\dots,\ell_b)$: a bounded
        contiguous span addressed by location.
\end{itemize}
The agent composes these in a closed loop: hypothesise a lexical anchor, search,
inspect hits against the outline, read a span wide enough to capture the
governing headers, and either extract the answer or refine. A trajectory is a
sequence $\tau=(o_1,r_1,\dots,o_T,r_T)$ terminating in an answer with cited
locations.

Crucially, \textsc{read} is span-addressed, not chunk-addressed: the agent
chooses $[a,b]$ \emph{after} seeing where its evidence lies, so it can extend
upward to include a unit header. Interface A cannot, because the partition was
fixed offline.

\paragraph{Determinism and auditability.}
Every $o_t$ is a pure function of $(D,\text{args})$, so re-executing $\tau$
reproduces the same results and any third party can replay the evidence chain.
This is not a promise about the design: we release the server and, for every
cell reported here, the exact line spans each system read, so the replay can be
performed against the reader's own copy of the document.
We call an answer \emph{grounded} if each figure it contains appears verbatim in
some span surfaced in $\tau$; this is mechanically checkable
(\Cref{sec:metrics}). Under Interface A the analogous check cannot localise: one
can verify the answer string occurs in the retrieved chunks, but not interrogate
why competing chunks were excluded.

\section{The Setting, Measured}
\label{sec:corpus}

\subsection{Testbed}

Our document is the \emph{Gujarat Finance Accounts 2024--25, Volume I}
\cite{gujarat2025finance}, the audited annual financial statement of an Indian
state government --- a public record, 780 printed pages, 3.97\,MB of PDF. It is representative of the class we
target: statutory, tabular, exactness-critical, and consumed by people who must
be able to check where a number came from.

We convert to Markdown with \texttt{pymupdf4llm}, selected over three
alternatives in a prior comparison for one reason: it is the only converter
tested that keeps a row label bound to its figures across columns. The
alternatives fail in ways that make any retrieval comparison meaningless ---
one collapses thirty table rows into a single cell, another severs row labels
from their numbers entirely. \Cref{app:conversion} records the comparison.

\subsection{What the Document Is}

\Cref{tab:corpus} reports the profile. Three quantities carry the argument.

\begin{table}[t]
\centering
\small
\begin{tabular}{lr}
\toprule
Property & Value \\
\midrule
Printed pages & 780 \\
Pages with recoverable folio & 745 \\
Lines (non-empty) & 19{,}198 \\
Characters & 1{,}683{,}730 \\
Statement / section headers & 204 \\
Table rows & 16{,}664 \\
\textbf{Table-row fraction} & \textbf{86.8\%} \\
Table columns (p50 / p90) & 7 / 10 \\
Numeric tokens & 58{,}791 \\
\textbf{Distinct numeric tokens} & \textbf{15{,}960} \\
\textbf{Numeric repetition ratio} & \textbf{3.68$\times$} \\
\bottomrule
\end{tabular}
\caption{Corpus profile. The document is overwhelmingly tabular, and its
numbers are heavily repeated: 58{,}791 numeric tokens take only 15{,}960
distinct values.}
\label{tab:corpus}
\end{table}

\textbf{It is a table, not prose.} 86.8\% of non-empty lines are table rows,
with a median of 7 columns. Retrieval methods tuned on prose benchmarks are
being applied here to a fundamentally different object.

\textbf{Its numbers are near-duplicates.} 58{,}791 numeric tokens collapse to
15{,}960 distinct values, a repetition ratio of 3.68. Thousands of similar
magnitudes, in similar row contexts, compete in a single embedding space. The
two operands of our motivating question --- $\rupee$\,48{,}964.89 crore and
$\rupee$\,18{,}942.94 crore --- sit in adjacent rows of the same statement.

\begin{figure}[t]
\centering
\includegraphics[width=0.82\linewidth]{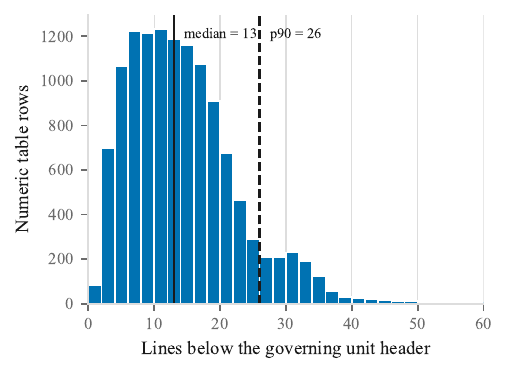}
\caption{Distance from a numeric table row to the header declaring its unit. A
chunk shorter than this distance cannot contain both, so the figures it carries
are ambiguous between lakh and crore --- a factor of 100.}
\label{fig:unit_distance}
\end{figure}

\textbf{Its units are inherited, not local.} This is the property we have not
seen reported, and it is the most consequential. The book declares units in
headers of the form ``($\rupee$ in lakh)'': 666 lines declare \emph{lakh} and 50
declare \emph{crore}. A figure inherits the unit of the nearest declaration
above it, at a distance of \textbf{median 13 lines, p90 26 lines, max 143
lines} (\Cref{fig:unit_distance}). Lakh and crore differ by a factor of 100. A
retrieval unit that does not contain the governing header therefore yields
figures that are ambiguous by two orders of magnitude --- and, unlike a missing
answer, this failure is invisible to the generator, which sees a plausible
number.

\subsection{What Chunking Destroys}

We make the consequence precise. For a chunk $c$ containing at least one numeric
table row, define $c$ \emph{unitless} if no line in $c$ declares a unit, and
\emph{yearless} if no line matches a fiscal-year column header. \Cref{tab:chunking}
reports both under two chunkers at matched size.

\begin{table}[t]
\centering
\small
\begin{tabular}{rrrrr}
\toprule
\multirow{2}{*}{Chunk} & \multicolumn{2}{c}{Fixed-size} & \multicolumn{2}{c}{Table-aware} \\
\cmidrule(lr){2-3}\cmidrule(lr){4-5}
 & Unitless & Yearless & Unitless & Yearless \\
\midrule
800   & 67.4\% & 73.8\% & 14.7\% & 30.1\% \\
1{,}500 & 36.3\% & 55.8\% & 1.0\%  & 28.7\% \\
2{,}000 & 18.0\% & 45.9\% & 0.3\%  & 28.9\% \\
2{,}500 & 11.6\% & 43.5\% & 0.2\%  & 28.3\% \\
4{,}000 & 5.4\%  & 40.5\% & 0.4\%  & \textbf{27.2\%} \\
\bottomrule
\end{tabular}
\caption{Context destroyed by chunking, over a 5$\times$ range of chunk size.
Fixed-size chunking degrades sharply as chunks shrink. Our table-aware chunker
carries the governing unit and fiscal-year rows across every boundary and
nearly eliminates unitless chunks --- but the yearless residue is
\emph{invariant} to chunk size, spanning only 2.9 points across the whole
sweep.}
\label{tab:chunking}
\end{table}

\begin{figure}[t]
\centering
\includegraphics[width=\linewidth]{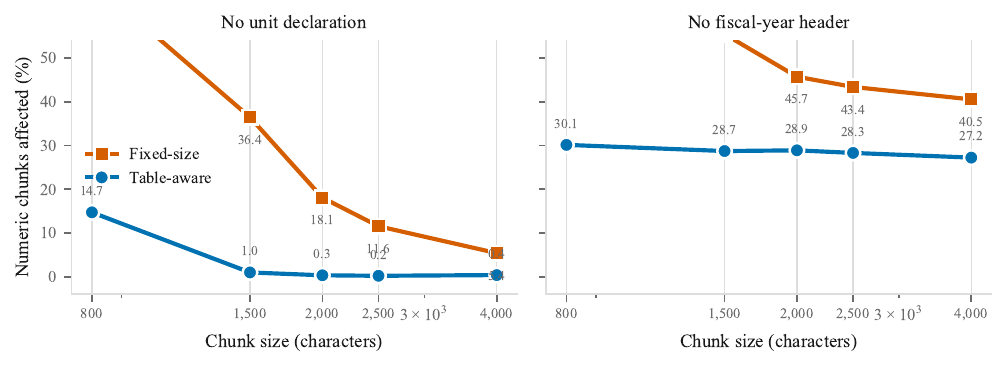}
\caption{The same data. Fixed-size chunking (squares) trades one failure for
another as size grows: fewer unitless chunks, but larger retrieval units.
Table-aware chunking (circles) removes the unit problem entirely and leaves the
year problem untouched at every size.}
\label{fig:chunk_damage}
\end{figure}

Three readings matter.

First, naive chunking degrades catastrophically at small chunk sizes: at 800
characters, \textbf{67.4\%} of numeric chunks carry no unit declaration. Smaller
chunks are usually recommended for precise retrieval; here they are precisely
what destroys the context that makes a figure interpretable.

Second, the fix works, and works cheaply. Our table-aware chunker searches
backwards up to 200 lines for the governing unit and year rows and prepends
them, subject to a budget so a carried header cannot consume the chunk. At
2{,}000 characters it reduces unitless chunks from 18.0\% to 0.3\%, a factor of
60.

Third --- and this is the load-bearing observation --- \textbf{the yearless
residue does not respond to tuning}. Across a five-fold range of chunk size it
moves between 27.2\% and 30.1\%, a spread of 2.9 points, with no trend that
tuning could exploit. Roughly 28\% of numeric chunks cannot be attributed to a
fiscal year no matter how the partition is drawn, because in this document the
year is declared once for a table that runs across many pages. A partition fixed
before the query is known has no way to recover it. An agent that reads a span
\emph{after} seeing where its evidence lies does.

We use the table-aware chunker for every retrieval baseline in this paper. A
baseline that fails for a fixable reason proves nothing, and we would rather
argue against the strongest version of the paradigm.

\begin{remark}
Chunk size is a soft target: a single converted table row can exceed 3{,}000
characters, and splitting it would sever the row. We report p95 chunk length,
which tracks the target closely, rather than the maximum, which does not.
\end{remark}

\section{\READ}
\label{sec:read}

\subsection{Implementation}

\READ{} --- Reliable Embedding-free Agentic Document-search --- is a read-only
MCP server \cite{anthropic2024mcp} over a directory of Markdown. It exposes the
three operation classes of \Cref{sec:interfaces} as four tools: \texttt{grep}
(normalized line search); \texttt{list} (files with sizes) and \texttt{outline}
(headings with line numbers), which together provide structural navigation; and
\texttt{read} (numbered lines, capped at 400 per call). Each is jailed to a root
directory and hard-capped so no single call can flood the context. There is no
embedding model, no vector index, and no learned component; the server's entire
state is the document text.

\subsection{Why Normalized Matching}
\label{sec:normalization}

A literal \texttt{grep} over converted PDF text is materially weaker than over
clean text, and this appears not to have been reported in prior agentic-search
work, which evaluates on already-clean corpora. Our converter introduces:
4{,}685 rows with a spurious leading empty column; 32{,}404 emphasis markers
welded to tokens; 1{,}719 words split across table cells; 1{,}037 repeated page
headers; and 24 distinct spellings of statement numbers across 654 mentions.
Most consequentially, 14{,}269 numbers are printed with digit grouping.

\READ's \texttt{grep} therefore matches against two normalized shadows of each
line --- one with emphasis and cell boundaries replaced by spaces, one with all
separators and thousands-commas removed --- while reporting original line
numbers, so results chain into \texttt{read}.

We measured which artifacts actually matter rather than assuming, probing twelve
realistic queries under normalized and literal matching. Normalization changes
the outcome for exactly two:

\begin{table*}[t]
\centering
\small
\begin{tabular}{llcc}
\toprule
Query & Kind & Norm. & Literal \\
\midrule
\texttt{48964.89} & figure, ungrouped & \checkmark & \ding{55} \\
\texttt{Statement Subject} & split across cells & \checkmark & \ding{55} \\
\midrule
\texttt{48,964.89} & figure, as printed & \checkmark & \checkmark \\
\texttt{Fiscal Deficit} & inside \texttt{**}\ldots\texttt{**} & \checkmark & \checkmark \\
\texttt{fiscal deficit} & lowercase & \checkmark & \checkmark \\
\texttt{Statement No.\ 17} & statement number & \checkmark & \checkmark \\
\texttt{Co-operation} & row label & \checkmark & \checkmark \\
\bottomrule
\end{tabular}
\caption{Matching sensitivity (7 of 12 probes shown). Normalization is
load-bearing for digit grouping and for words split across table cells. It is
\emph{not} needed for case, for emphasis wrapping a whole phrase, or for
statement-number punctuation --- we had assumed the latter mattered and it does
not.}
\label{tab:normalization}
\end{table*}

The first row is the important one. On this document the agent's queries
\emph{are} figures --- reverse lookups, cross-checks, verification of a computed
value --- and an agent that has derived or recalled a number writes it without
separators. Digit grouping is precisely what breaks that search, and 14{,}269 of
the document's numbers carry it. We report this narrower mechanism rather than
the broader one we first assumed, and we retain \texttt{read\_naive\_grep} as an
ablation that isolates it.


\subsection{The Conversion Ceiling}
\label{sec:conversion}

Every retrieval comparison on a converted PDF silently assumes the text is a
faithful rendering of the document. It is not, and the gap is neither uniform
noise nor negligible: it is structured, it is measurable, and where it bites it
changes a figure by two orders of magnitude.

The clearest case is Statement No.\ 18, which lists loans sanctioned in
perpetuity. The printed interest rates are 11.00 and 10.00 per cent. The
conversion places the digits in one table cell and the decimal point in the cell
\emph{below}:

\begin{quote}\small\ttfamily
16458: ...|500.00|**1100**| \\
16459: ...|471.50|**.**| \\
16460: ...|500.00|**1000**| \\
16461: ...|500.00|**.**|
\end{quote}

An agent that retrieves these lines perfectly still reads 1100 where the
document prints 11.00. The same pattern appears in Statement No.\ 2, where the
Consolidated Fund deficit \rupee\,26{,}837.65 crore is split into the digit run
\texttt{2683765} on one line and the separators \texttt{,.} on the next, so the
well-formed figure occurs nowhere in the corpus.

\begin{table}[t]
\centering
\small
\begin{tabular}{lr}
\toprule
Class & Count \\
\midrule
Suspect bare integer (missing decimal) & 174 \\
Orphaned separator cell & 139 \\
Word split across cells & 133 \\
Detached separator (owner identified) & 22 \\
Decimal detached from its digits & 15 \\
\midrule
Total instances & 483 \\
Table rows affected & 440 (2.6\%) \\
\midrule
Numeric rows affected & 381 (3.0\%) \\
\textbf{Numeric rows with a magnitude error} & \textbf{11 (0.09\%)} \\
\bottomrule
\end{tabular}
\caption{Conversion damage in the corpus. The book's nil convention (``..'',
17{,}734 cells) is excluded --- counting it as damage overstates the total by
roughly 40$\times$, an error we made on a first pass. Damage is rare but
concentrated: the 15 detached decimals sit almost entirely in two statements.}
\label{tab:conversion}
\end{table}

\Cref{tab:conversion} quantifies this across the document. Three points follow.

\textbf{It is rare but concentrated.} Only 3.0\% of numeric rows are affected
and 0.09\% carry an outright magnitude error, so this is not a pervasive
degradation. But the affected rows cluster: a question that happens to target
Statement No.\ 18 is far more likely to be unanswerable-as-converted than the
aggregate rate suggests.

\textbf{It is a ceiling, not a confound.} No retriever --- dense, sparse, or
agentic --- can recover a decimal point that is in a different cell from its
digits. Accuracy measured on such a question is measuring the converter, not the
retrieval interface. We therefore give these questions their own category
(\texttt{conversion\_limited}) rather than charging their failures to whichever
system was under test, and we report them separately.

\textbf{It reframes what a retrieval failure means.} In an informal evaluation
of this document, a system was judged to have failed on the Statement No.\ 18
question. Inspection showed it had retrieved exactly the right lines and read
them faithfully; the source text was wrong. Distinguishing ``did not find the
evidence'' from ``found evidence that the conversion had already destroyed''
requires the trajectory, which is available for \READ{} and for top-$k$ systems
alike but not for a system that reports only an answer.

A second observation from the same evaluation is worth recording because it is
\emph{not} a failure. Statement No.\ 18 is denominated in lakh while most
aggregate statements are in crore, so the same amount is correctly reported as
either \rupee\,10{,}597.61 lakh or \rupee\,105.98 crore. A grader that does not
normalise units scores one of these wrong. Our numeric matcher normalises lakh,
crore, and thousand before comparison for exactly this reason
(\Cref{sec:metrics}).

\section{Evaluation Protocol}
\label{sec:protocol}

\subsection{Systems}

All systems share one backbone, one answer format, and one accounting of tokens,
cost, and latency, so a difference between them is a difference of interface.

\begin{table}[t]
\centering
\small
\begin{tabular}{lp{4.4cm}}
\toprule
System & Description \\
\midrule
\READ & Agent + list/outline/grep/read \\
\READ$_{\text{lit}}$ & Same, literal grep (ablation) \\
\READ$_{-\text{out}}$ & grep + read only (ablation) \\
\midrule
Dense & Chunk $\to$ embed $\to$ top-$k$ \\
BM25 & Sparse top-$k$ \\
Hybrid & RRF(dense, BM25) + cross-encoder \\
\midrule
AgenticVec & Agent + vector-search tool \\
Oracle & Chunks overlapping gold lines \\
LongCtx & Whole document in context \\
\bottomrule
\end{tabular}
\caption{Systems under comparison. \READ$_{\text{lit}}$ isolates normalized
matching; AgenticVec isolates interface from iteration.}
\label{tab:systems}
\end{table}

Two entries carry the argument. \READ$_{\text{lit}}$ isolates normalized
matching (\Cref{sec:normalization}). \textbf{AgenticVec} is the critical
control: an agent with the same loop and budget but a vector-search tool instead
of \READ's operations. If \READ's advantage came merely from being able to
iterate, AgenticVec would close the gap; if it comes from the interface, it will
not. Without this control, any result is attributable to iteration alone.

Retrieval baselines use the table-aware chunker (\Cref{tab:chunking}) and a
local \texttt{bge-base-en-v1.5} encoder. Embeddings run locally so the baseline
does not drift when a hosted embedding endpoint updates
\cite{chen2023chatgptdrift}; \Cref{app:sweep} reports the chunk-size and $k$
sweep, and we report the best dense configuration found rather than the first.

\subsection{Benchmark}

Questions span six categories: single-figure lookup, multi-row aggregation,
cross-statement arithmetic, structural navigation, unanswerable probes, and
\emph{conversion-limited} items whose answer the converter destroyed
(\Cref{sec:conversion}).
Gold answers are transcribed from the document by the authors and then checked
mechanically: every expected figure must appear within one line of its declared
gold line, and the column mapping of each source table is verified against an
arithmetic identity the table itself satisfies rather than against our reading
of its header (\Cref{app:data}). Arithmetic questions carry both the derived
answer and its operands, and a validator asserts that the derived value
\emph{does not} appear anywhere in the corpus --- otherwise the question is a
lookup in disguise. No model generates or grades a gold answer.

The unanswerable probes include \emph{false-premise} items, in which the
figures exist but not the quantity the question presupposes --- for instance
asking for a rate of interest from a statement that reports only interest
amounts. These detect a failure mode accuracy alone cannot: a system that
accepts the premise and fabricates a matching citation. We have no clean
instance of that failure in our data; the one candidate in the pilot turned out
to be an artifact of our own analysis code (\Cref{sec:pilot}).

\subsection{Metrics}
\label{sec:metrics}

Three independent signals, because no one of them is trustworthy alone.

\textbf{Exact-figure accuracy.} Deterministic numeric comparison after
normalizing Indian digit grouping, parenthesised and \texttt{(-)} negatives, and
lakh/crore units, at 0.5\% relative tolerance --- enough to equate
$48{,}964.89$ and $48{,}964.9$ while separating $48{,}964$ from $18{,}942$.

\textbf{LLM judge.} A grader with the reference answer in hand, held fixed
across systems, emitting a structured verdict. We report agreement with the
deterministic signal; disagreements are read by hand.

\textbf{Groundedness.} Does every figure in the answer appear in text the system
actually retrieved or read? This is purely mechanical, and it is the metric a
fluent-but-unsupported answer cannot pass. It is computable for \READ{} and for
top-$k$ systems alike, since both expose exactly what they surfaced.

We additionally report hallucination rate on unanswerable probes, page-level
evidence coverage, tool-call counts, tokens, cost, and latency. Comparisons are
paired: exact McNemar on discordant pairs, bootstrap confidence intervals on the
delta, Holm-corrected across the family of baselines. We report the benchmark's
minimum detectable difference so that a null result is reported as underpowered
rather than as no difference.

\section{Pilot Observations}
\label{sec:pilot}

\begin{table}[t]
\centering
\small
\begin{tabular}{llrrl}
\toprule
Backbone & System & Turns & In-tok & Outcome \\
\midrule
\multirow{2}{*}{2.5-pro} & \READ & 4 & 9{,}428 & correct \\
 & Dense & 1 & 6{,}702 & abstained \\
\midrule
\multirow{2}{*}{3.6-flash} & \READ & 7 & 58{,}320 & correct$^\dagger$ \\
 & Dense & 1 & 6{,}702 & abstained \\
\bottomrule
\end{tabular}
\caption{Pilot on the motivating question, two backbones. $^\dagger$We first
recorded this cell as citing the wrong statement. That was our own parsing bug,
not the model's error, and the retraction is described below.
\textbf{This is $n{=}2$ and is not a result}; it is reported to show the failure
mode concretely.}
\label{tab:pilot}
\end{table}

Before the full evaluation we ran a two-question pilot to exercise the harness.
We report it because it makes the mechanism concrete, and we state plainly that
$n{=}2$ supports no quantitative claim.

The dense baseline abstained on every cell, reporting that the Fiscal Deficit
was not present in its retrieved excerpts. The two operands lie on lines 385 and
432 --- 47 lines apart, in the same statement --- and no top-8 slice of
2{,}000-character chunks contained both. \READ{} located both and computed the
difference, $\rupee$\,30{,}021.95 crore, a value that appears nowhere in the
document.

Two further observations shaped the protocol. First, \emph{iteration is
expensive}: \READ{} consumed 58--78k input tokens against the dense baseline's
6.7--8.2k, roughly an order of magnitude, and any cost-accuracy claim must carry
this.

Second, a methodological caution we record because it nearly became a result. We
initially read one backbone's answer as citing the wrong statement --- a
provenance failure invisible to exact-figure accuracy. On inspection the model
was right and our analysis was wrong: our section-attribution code recognised
only Markdown headings, while the converter renders each statement's running
header as a \emph{table row} with the title split across cells. Roughly a
hundred lines were therefore attributed to the preceding statement. The lesson
is not about the model. It is that a parsing bug in the analysis layer can
manufacture a confident, plausible finding about model behaviour, and that such
a finding is most likely to survive review when it is the one the authors
expected to see. We caught it by reading the raw lines, not by any metric.

\section{Results}
\label{sec:results}

All systems use \texttt{gemini-2.5-pro} and see the same 51 questions. Every
comparison is paired on the question, tested with an exact McNemar test, and
Holm-corrected across the family of six baselines. Confidence intervals on
accuracy are Wilson; intervals on differences are paired bootstrap.

\subsection{Main comparison}

\begin{table*}[t]
\centering
\small
\caption{Accuracy and behaviour over 51 questions. \emph{Grounded} is the
fraction of answers in which every figure stated appears in text the system
actually retrieved or read. \emph{Halluc.} is the fraction of the five
unanswerable probes answered rather than declined. Intervals are Wilson 95\%.}
\label{tab:main}
\begin{tabular}{lccccc}
\toprule
system & acc.\ (\%) & 95\% CI & grounded & halluc. & \$/q \\
\midrule
\multicolumn{6}{l}{\emph{\READ{} and its ablations}} \\
\quad no-outline   & 66.7 & [53.0,\,78.0] & 66.7 & 40 & 0.040 \\
\quad naive-grep   & 66.7 & [53.0,\,78.0] & 60.4 &  0 & 0.043 \\
\quad \textbf{\READ} & \textbf{58.8} & [45.2,\,71.2] & 58.0 & 40 & 0.058 \\
\midrule
\multicolumn{6}{l}{\emph{retrieval baselines}} \\
\quad BM25         & 51.0 & [37.7,\,64.1] & 67.3 & 40 & 0.020 \\
\quad hybrid       & 29.4 & [18.7,\,43.0] & 72.0 & 60 & 0.022 \\
\quad agentic RAG  & 27.5 & [17.1,\,40.9] & 42.6 &  0 & 0.050 \\
\quad dense        & 15.7 & [\phantom{0}8.2,\,28.0] & 47.9 & 80 & 0.023 \\
\bottomrule
\end{tabular}
\end{table*}

\begin{table}[t]
\centering
\small
\caption{Paired comparisons against \READ. Positive $\Delta$ favours \READ.
$p$ is exact McNemar, Holm-corrected over the six comparisons; $\ast$ marks
$p_{\text{Holm}}<0.05$. \emph{dense} here
is the configuration of \Cref{sec:protocol} (2{,}000-character table-aware
chunks, $k{=}8$); \Cref{app:sweep} sweeps that configuration and reports the
comparison against dense's best setting.}
\label{tab:paired}
\begin{tabular}{lrcrl}
\toprule
vs. & $\Delta$ (pp) & 95\% CI & $p_{\text{Holm}}$ & \\
\midrule
dense        & $+43.1$ & [27.5,\,56.9] & 0.00002 & $\ast$ \\
agentic RAG  & $+31.4$ & [13.7,\,49.0] & 0.012   & $\ast$ \\
hybrid       & $+29.4$ & [11.8,\,47.1] & 0.016   & $\ast$ \\
BM25         & $+7.8$  & [$-9.8$,\,25.5] & 1.00  &        \\
naive-grep   & $-7.8$  & [$-23.5$,\,7.8] & 1.00  &        \\
no-outline   & $-7.8$  & [$-21.6$,\,5.9] & 1.00  &        \\
\bottomrule
\end{tabular}
\end{table}

\READ{} answers 58.8\% of the benchmark against dense retrieval's 15.7\%, a
paired difference of 43.1 points that survives correction at
$p_{\text{Holm}}=2\times10^{-5}$ (\Cref{tab:paired}). The same holds against the
hybrid pipeline ($+29.4$, $p=0.016$).

The comparison we consider most informative is \emph{agentic RAG}: the same
agent, the same backbone, the same turn budget, differing only in that its
retrieval tool returns top-$k$ chunks instead of \READ's four operations. It
scores 27.5\%. Iteration alone therefore does not close the gap --- an agent
allowed to query a vector index repeatedly remains 31.4 points behind the same
agent given a lexical, structural interface ($p=0.012$). The gain belongs to the
interface, not to the loop.

\subsection{What does not support a stronger claim}

\textbf{BM25 is not beaten.} At 51.0\% it sits 7.8 points below \READ{} with a
confidence interval spanning zero ($p=1.00$). With 51 questions the smallest
difference this design can resolve at 80\% power is 27.7 points, so this is a
null result, not evidence of equivalence. That threshold is an \emph{a priori},
unpaired approximation at a base rate of one half; the paired test is more
sensitive, because it conditions on discordant pairs alone --- in
\Cref{app:sweep} a 23.5-point gap does reach significance. It is therefore a
guide to what this benchmark can be expected to see, not a decision rule. What
makes the BM25 comparison uninformative is more specific: of 22 discordant
questions, 13 favour \READ{} and 9 favour BM25, which is close to a coin toss.
The honest reading is that our evidence
separates \emph{embedding-based} retrieval from \emph{embedding-free}
retrieval, and does not separate the two embedding-free systems from each
other. That is a narrower claim than ``agentic search wins'', and it happens to
be the claim the paper set out to test: on this document, the vector index is
not merely unnecessary --- it is the component doing the damage. BM25 also costs
a third as much as \READ{} and runs in a third the latency, which matters for
anyone choosing a system rather than testing a hypothesis.

\textbf{Both ablations score above \READ.} Removing \texttt{memory\_outline},
or replacing normalized matching with raw regular expressions, each yields
66.7\% against \READ's 58.8\%. Neither difference approaches significance
($p=1.00$), and at $n=51$ a 7.8-point gap is four questions. We investigated
rather than dismissed it (\Cref{sec:pilot}): the six questions driving the
inversion decompose into one harness failure, one genuine sign error, one
correct answer buried in hedging, and three real search failures. We therefore
cannot claim the full operation set is the best configuration --- only that it
is not detectably worse. A larger benchmark may well show that outline access
is dead weight on a document whose headings the converter mangles.

\textbf{\READ{} is not the most grounded system.} At 58.0\% it trails BM25
(67.3\%) and hybrid (72.0\%). A system that reads more text has more
opportunity to state a figure that its own retrieved text does not contain, and
\READ{} exposes far more lines per question than a top-$k$ retriever does. This
cuts against the interpretability argument and we report it as such.

\textbf{The dense baseline can be tuned well above the figure we report.}
\Cref{tab:main} lists dense at 15.7\%, which is its accuracy at $k{=}8$ --- the
depth every retrieval baseline in this paper uses, so that the comparison is
matched. It is not dense's best. Sweeping chunk size and depth
(\Cref{app:sweep}) finds 35.3\% at 2{,}000 characters and $k{=}16$, more than
double. The mechanism is the one \Cref{sec:corpus} predicts: a top-$k$ system
cannot ensure a chunk arrives with the header that gives its figures a unit and
a year, so its only recourse is to retrieve more chunks and cover the header by
chance. Doubling $k$ buys roughly that. We therefore do not claim dense
retrieval scores 15.7\% on this document --- we claim it scores 35.3\% when
tuned, at 41\% higher cost per question, and that \READ{} still beats that
configuration by 23.5 points ($p_{\text{Holm}}=0.017$). Every one of the six
configurations we swept is beaten significantly.

\subsection{Where the difference comes from}

\begin{table*}[t]
\centering
\small
\caption{Accuracy (\%) by category. \emph{Conv.-lim.} questions have evidence
damaged in PDF conversion (\Cref{sec:conversion}); no system can be expected to
reach 100\%.}
\label{tab:category}
\begin{tabular}{lcccccc}
\toprule
system & look. & arith. & agg. & nav. & conv.-lim. & unans. \\
       & (24)  & (9)    & (5)  & (6)  & (2)        & (5)    \\
\midrule
no-outline  & 70.8 & 33.3 & 100.0 & 83.3 & 50.0 & 60.0 \\
naive-grep  & 75.0 & 33.3 & 60.0  & 66.7 & 50.0 & 100.0 \\
\READ       & 70.8 & 44.4 & 40.0  & 50.0 & 50.0 & 60.0 \\
BM25        & 62.5 & 33.3 & 20.0  & 66.7 & 0.0  & 60.0 \\
hybrid      & 20.8 & 0.0  & 60.0  & 66.7 & 50.0 & 40.0 \\
agentic RAG & 16.7 & 11.1 & 20.0  & 50.0 & 0.0  & 100.0 \\
dense       & 12.5 & 11.1 & 0.0   & 50.0 & 0.0  & 20.0 \\
\bottomrule
\end{tabular}
\end{table*}

\begin{figure*}[t]
\centering
\includegraphics[width=\linewidth]{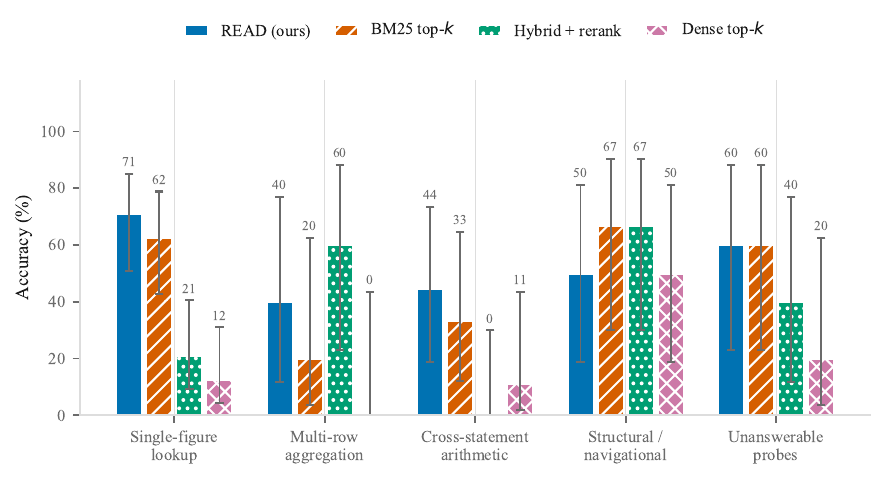}
\caption{Accuracy by question category, \READ{} against the three retrieval
baselines. Bars are Wilson 95\% intervals; the categories carry 24, 5, 9, 6 and
5 questions respectively, so the intervals are wide and most single-category
differences are not individually significant. The pattern that matters is the
first group: on single-figure lookup, where the document is densest in
near-identical numbers, dense retrieval reaches 12\% and \READ{} 71\%. On
structural questions, which are carried by prose rather than digits, the four
systems are indistinguishable.}
\label{fig:main}
\end{figure*}

The gap is concentrated in \emph{lookup}, the largest category: 70.8\% for
\READ{} against 12.5\% for dense (\Cref{tab:category}, \Cref{fig:main}). These are the questions
that ask for one printed figure from a named row and column, which is precisely
where thousands of near-identical numbers compete in one embedding space
(\Cref{sec:corpus}). On \emph{navigation} --- questions about which statement,
which page, which unit --- every system does comparatively well, because the
answer is carried by prose and headings rather than by digits. Dense retrieval
is not uniformly weak; it is weak exactly where the document is numeric.

\begin{table*}[t]
\centering
\small
\caption{Failure taxonomy: counts of failing cells by cause, over 51 questions.
\emph{Never surfaced} means the gold evidence never entered the model's context.}
\label{tab:failures}
\begin{tabular}{lccc}
\toprule
system & never surfaced & abstained (answerable) & ungrounded \\
\midrule
\READ       &  4 &  1 & 11 \\
naive-grep  &  6 &  0 &  9 \\
no-outline  &  5 &  0 &  7 \\
BM25        &  7 &  7 &  8 \\
dense       & 13 & 11 & 14 \\
hybrid      & 15 & 12 &  5 \\
agentic RAG &  5 &  8 & 21 \\
\bottomrule
\end{tabular}
\end{table*}

\Cref{tab:failures} isolates the mechanism. Dense retrieval fails to place the
gold evidence in context on 13 questions and hybrid on 15, against \READ's 4.
The consequence is visible in the next column: dense declines to answer 11
answerable questions and hybrid 12, against \READ's 1. This is the failure mode
from \Cref{sec:intro} --- retrieval declining to deliver --- and it is the
common case rather than a curiosity. Across the whole benchmark, dense
retrieval surfaces a gold page in only 4.3\% of questions, against \READ's
41.3\%.

Two further observations. First, on the two \emph{conversion-limited} questions
only the \READ{} family and hybrid score at all; the purely lexical and purely
dense retrievers score zero. Second, agentic RAG shows the opposite profile to
dense: it rarely fails to surface \emph{something} (5 never-surfaced) but
produces 21 ungrounded answers, the worst of any system --- an agent that keeps
querying a vector index accumulates plausible context and then reasons over it
past the point the evidence supports.

\subsection{Behaviour on unanswerable questions}

Five questions have no answer in the document. Dense retrieval answers four of
them anyway (80\%), hybrid three, \READ{} and BM25 two. The interesting pair is
the two systems that never fabricated: naive-grep and agentic RAG, both at 0\%.
For agentic RAG this is not a virtue --- it also abstains on 8 answerable
questions, so it is simply a reticent system. With only five probes, none of
these differences is individually meaningful; we report the counts rather than
rates for that reason, and treat the pattern as suggestive at best.

\subsection{Cost}

\READ{} is the most expensive system per question (\$0.058 against dense's
\$0.023) and the slowest (31.4\,s against 12.7\,s), because it makes 4.8 tool
calls per question and reads substantially more text. Against BM25 --- the
baseline it does not significantly beat --- it costs 2.9$\times$ as much for
7.8 points that may be noise. A practitioner reading only \Cref{tab:main} should
take BM25 seriously. The case for \READ{} rests on the failure taxonomy and on
auditability, not on the cost--accuracy frontier.

\section{Limitations}
\label{sec:limitations}

\textbf{One document, one backbone, 51 questions.} Every number here comes from
a single financial report read by a single model. Nothing establishes that the
result transfers to another document class, another accounting convention, or a
weaker backbone --- \READ{} presumes a model that can plan a search, and we have
not measured where that ability gives out. At this size the benchmark is
expected to resolve paired differences of roughly 28 points, so the 7.8-point
\READ--BM25 gap is unresolved rather than absent.

\textbf{The benchmark was built by the authors of the system under test.} Gold
answers are validated mechanically against the document rather than by the
drafter's own reading, and the categories where \READ{} is weakest are reported
rather than dropped (\Cref{app:data}). Neither of those removes the conflict.

\textbf{The benchmark grew while the experiment ran, and the growth was not
blind.} An earlier 25-question version produced no significant difference
between any pair of systems. Two further batches were added after inspection
revealed coverage defects --- first that no question exercised column selection
in a wide table, then that 31 of 36 questions drew their evidence from printed
pages 2--18 of a 745-page volume. Both defects were identified from the
document rather than from any system's results, and questions were drafted
without running them first. But both happen to fall where chunk-based retrieval
is weak, and we knew that when we chose to fix them. A reader should treat the
$n{=}51$ result as one we sought and found, not one we stumbled into; the
pre-registered version of this study would have fixed the benchmark first. We
report the accuracy of dense retrieval at each stage --- 24.0\%, 16.7\%, 15.7\%
--- so the drift is visible rather than hidden.

\textbf{\READ{} is not the most grounded system}, trailing BM25 and hybrid on
the fraction of answers whose every figure appears in retrieved text
(\Cref{sec:results}). An interpretability argument that cannot demonstrate
superior groundedness is weaker for it.

\textbf{Scope.} Claims are limited to single, long, structured documents
consumed by capable tool-calling models. Lexical anchoring presumes the answer's
vocabulary is guessable from the question; where queries and evidence share
little surface vocabulary, dense retrieval's paraphrase tolerance may retain an
edge and hybrid designs remain sensible.

\textbf{One document, one converter.} All measurements come from a single report
converted by a single tool. The conversion ceiling in \Cref{sec:conversion} is a
property of \texttt{pymupdf4llm} on this PDF, not a universal constant; a
different converter would move it, and a commercial layout model would likely
lower it at a cost we did not pay. What we claim is that the ceiling exists and
must be measured, not that this particular value generalises. The unit-inheritance property is common to Indian
government financial statements but its distances are specific to this document;
the artifact counts are specific to \texttt{pymupdf4llm}. Generality requires
more documents, more converters, and other domains.

\textbf{Cost.} \READ's per-query cost is variable and can exceed a fixed top-$k$
budget substantially, as the pilot shows. \READ{} trades worst-case cost for
correctness and auditability, which is the right trade in high-stakes settings
and not obviously the right one in high-volume consumer search.

\textbf{What this should not be read as claiming.} Not that embeddings are
obsolete, and not that dense retrieval is weak in general --- BEIR-style
benchmarks measure something real. The claim is narrower: for documents whose
meaning depends on inherited layout context, a partition fixed before the query
is the wrong primitive.

\section{Conclusion}
\label{sec:conclusion}

We challenged the assumption that question answering over long documents
requires embedding-based retrieval, and we grounded the challenge in
measurement. On a 780-page financial report, 86.8\% of content is tabular,
numbers repeat 3.68-fold, and units are inherited from headers a median of 13
lines above the figures they govern. Fixed-size chunking leaves up to 67.4\% of
numeric chunks unitless; a table-aware chunker built as a steelman reduces that
to 0.3\%, but 27--30\% still lack a fiscal-year header at every chunk size
tested. That residue is flat across a five-fold range of chunk size: it is not a
tuning failure but what happens when a partition is fixed before the query is
known.

\READ{} replaces the chunk--embed--index pipeline with deterministic
operations, turning retrieval from an opaque ranking into a replayable
procedure. On 51 questions it answers 58.8\% against dense retrieval's 15.7\%,
and the mechanism behind the gap is visible rather than inferred: dense
retrieval never places the required evidence in context on 13 questions, where
\READ{} fails to on 4. An agent handed the same loop and backbone but a top-$k$
tool reaches 27.5\%, so the gain is attributable to the interface and not to
iteration.

We are equally clear about the boundary of the claim. A BM25 baseline is
statistically indistinguishable from \READ{} at a third of the cost. What the
evidence separates is embedding-based from embedding-free retrieval on this
document --- not agentic search from lexical search. The vector index was not
merely unnecessary here; it was the component doing the damage, and a
practitioner with a lexical index and no agent would already have captured most
of the available gain.

As models become competent agents, the retrieval question shifts from
\emph{which retriever ranks best} to \emph{which interface lets the agent see,
verify, and account for the evidence}. For long structured documents we argue
the answer is the simplest one: let the model read.

\section*{Code and Data Availability}

Everything needed to check or extend this work is at
\url{https://github.com/twospoon/READ}, under the MIT licence: the \READ{} MCP
server, the four converter backends, the full evaluation harness, the benchmark,
and the raw result records for all 663 cells reported here.

Two properties of the release are deliberate. First, every table and figure in
this paper can be re-derived from the shipped result records \emph{without any
model calls} --- an API key is needed only to generate new rows, so verifying our
arithmetic costs nothing. Second, the records store line numbers, spans, and
counts but no document text, which is what makes the auditability claim of
\Cref{sec:interfaces} checkable: the \texttt{exposed\_lines} field of any cell,
replayed against the reader's own conversion, reconstructs exactly what that
system saw.

We do not redistribute the source document or its Markdown conversion. The
benchmark ships questions, hand-read gold answers, and line references; the
repository gives the two commands that fetch the report from the Comptroller and
Auditor General of India and convert it, together with a verifier that reports
whether a local conversion has drifted from the gold line numbers.

\section*{Acknowledgements}

We thank Two Spoon Technologies Private Limited (TwoSpoon) for sponsoring this
research and for funding the model calls the experiments required. All three
authors are affiliated with TwoSpoon, and the work was carried out with no
external funding.

\paragraph{Author contributions.}
S.T.\ led the research: the question, the corpus analysis, the \READ{} interface
and its evaluation harness, the benchmark, and the experiments reported here.
T.H.\ contributed to the experimental work. A.V.\ conducted the code review and
the evaluation audit, and prepared the final manuscript.

\bibliography{main_paper}

\begin{thebibliography}{28}
\providecommand{\natexlab}[1]{#1}
\providecommand{\url}[1]{\texttt{#1}}
\expandafter\ifx\csname urlstyle\endcsname\relax
  \providecommand{\doi}[1]{doi: #1}\else
  \providecommand{\doi}{doi: \begingroup \urlstyle{rm}\Url}\fi

\bibitem[{Anthropic}(2024)]{anthropic2024mcp}
{Anthropic}.
\newblock Introducing the {Model Context Protocol}.
\newblock \url{https://www.anthropic.com/news/model-context-protocol}, 2024.
\newblock Accessed 2026-08-03.

\bibitem[Asai et~al.(2020)Asai, Hashimoto, Hajishirzi, Socher, and
  Xiong]{asai2020path}
Asai, A., Hashimoto, K., Hajishirzi, H., Socher, R., and Xiong, C.
\newblock Learning to retrieve reasoning paths over wikipedia graph for
  question answering, 2020.
\newblock URL \url{https://arxiv.org/abs/1911.10470}.

\bibitem[Asai et~al.(2023)Asai, Wu, Wang, Sil, and Hajishirzi]{asai2023selfrag}
Asai, A., Wu, Z., Wang, Y., Sil, A., and Hajishirzi, H.
\newblock Self-rag: Learning to retrieve, generate, and critique through
  self-reflection, 2023.
\newblock URL \url{https://arxiv.org/abs/2310.11511}.

\bibitem[Chen et~al.(2024)Chen, Zaharia, and Zou]{chen2023chatgptdrift}
Chen, L., Zaharia, M., and Zou, J.
\newblock How is {ChatGPT}'s behavior changing over time?
\newblock \emph{Harvard Data Science Review}, 6\penalty0 (2), mar 12 2024.
\newblock https://hdsr.mitpress.mit.edu/pub/y95zitmz.

\bibitem[Chen et~al.(2021)Chen, Tworek, Jun, Yuan, de~Oliveira~Pinto, Kaplan,
  Edwards, Burda, Joseph, Brockman, Ray, Puri, Krueger, Petrov, Khlaaf, Sastry,
  Mishkin, Chan, Gray, Ryder, Pavlov, Power, Kaiser, Bavarian, Winter, Tillet,
  Such, Cummings, Plappert, Chantzis, Barnes, Herbert-Voss, Guss, Nichol,
  Paino, Tezak, Tang, Babuschkin, Balaji, Jain, Saunders, Hesse, Carr, Leike,
  Achiam, Misra, Morikawa, Radford, Knight, Brundage, Murati, Mayer, Welinder,
  McGrew, Amodei, McCandlish, Sutskever, and Zaremba]{chen2021codex}
Chen, M., Tworek, J., Jun, H., Yuan, Q., de~Oliveira~Pinto, H.~P., Kaplan, J.,
  Edwards, H., Burda, Y., Joseph, N., Brockman, G., Ray, A., Puri, R., Krueger,
  G., Petrov, M., Khlaaf, H., Sastry, G., Mishkin, P., Chan, B., Gray, S.,
  Ryder, N., Pavlov, M., Power, A., Kaiser, L., Bavarian, M., Winter, C.,
  Tillet, P., Such, F.~P., Cummings, D., Plappert, M., Chantzis, F., Barnes,
  E., Herbert-Voss, A., Guss, W.~H., Nichol, A., Paino, A., Tezak, N., Tang,
  J., Babuschkin, I., Balaji, S., Jain, S., Saunders, W., Hesse, C., Carr,
  A.~N., Leike, J., Achiam, J., Misra, V., Morikawa, E., Radford, A., Knight,
  M., Brundage, M., Murati, M., Mayer, K., Welinder, P., McGrew, B., Amodei,
  D., McCandlish, S., Sutskever, I., and Zaremba, W.
\newblock Evaluating large language models trained on code, 2021.
\newblock URL \url{https://arxiv.org/abs/2107.03374}.

\bibitem[{Comptroller and Auditor General of India}(2025)]{gujarat2025finance}
{Comptroller and Auditor General of India}.
\newblock Finance accounts 2024--25, volume i.
\newblock Office of the Principal Accountant General (A\&E), Gujarat, 2025.
\newblock URL
  \url{https://cag.gov.in/uploads/state_accounts_report/account-report-FA-VOL-I-2024-25-069c52aa2b34bf9-63690940.pdf}.

\bibitem[Cormack et~al.(2009)Cormack, Clarke, and Buettcher]{cormack2009rrf}
Cormack, G.~V., Clarke, C. L.~A., and Buettcher, S.
\newblock Reciprocal rank fusion outperforms condorcet and individual rank
  learning methods.
\newblock In \emph{Proceedings of the 32nd International ACM SIGIR Conference
  on Research and Development in Information Retrieval}, SIGIR '09, pp.\
  758--759, New York, NY, USA, 2009. Association for Computing Machinery.
\newblock ISBN 9781605584836.
\newblock \doi{10.1145/1571941.1572114}.
\newblock URL \url{https://doi.org/10.1145/1571941.1572114}.

\bibitem[Dong et~al.(2024)Dong, Zhao, Tian, Xiong, Zhou, Lin, Cambronero, He,
  Han, and Zhang]{dong2024spreadsheetllm}
Dong, H., Zhao, J., Tian, Y., Xiong, J., Zhou, M., Lin, Y., Cambronero, J., He,
  Y., Han, S., and Zhang, D.
\newblock Encoding spreadsheets for large language models.
\newblock In Al-Onaizan, Y., Bansal, M., and Chen, Y.-N. (eds.),
  \emph{Proceedings of the 2024 Conference on Empirical Methods in Natural
  Language Processing}, pp.\  20728--20748, Miami, Florida, USA, November 2024.
  Association for Computational Linguistics.
\newblock \doi{10.18653/v1/2024.emnlp-main.1154}.
\newblock URL \url{https://aclanthology.org/2024.emnlp-main.1154/}.

\bibitem[Formal et~al.(2021)Formal, Piwowarski, and
  Clinchant]{formal2021splade}
Formal, T., Piwowarski, B., and Clinchant, S.
\newblock Splade: Sparse lexical and expansion model for first stage ranking.
\newblock In \emph{Proceedings of the 44th International ACM SIGIR Conference
  on Research and Development in Information Retrieval}, SIGIR '21, pp.\
  2288--2292, New York, NY, USA, 2021. Association for Computing Machinery.
\newblock ISBN 9781450380379.
\newblock \doi{10.1145/3404835.3463098}.
\newblock URL \url{https://doi.org/10.1145/3404835.3463098}.

\bibitem[Gao et~al.(2024)Gao, Xiong, Gao, Jia, Pan, Bi, Dai, Sun, Wang, and
  Wang]{gao2023ragsurvey}
Gao, Y., Xiong, Y., Gao, X., Jia, K., Pan, J., Bi, Y., Dai, Y., Sun, J., Wang,
  M., and Wang, H.
\newblock Retrieval-augmented generation for large language models: A survey,
  2024.
\newblock URL \url{https://arxiv.org/abs/2312.10997}.

\bibitem[Gulati et~al.(2026)Gulati, Sen, Sarguroh, and Paul]{gulati2026frtr}
Gulati, A., Sen, S., Sarguroh, W., and Paul, K.
\newblock From rows to reasoning: A retrieval-augmented multimodal framework
  for spreadsheet understanding, 2026.
\newblock URL \url{https://arxiv.org/abs/2601.08741}.

\bibitem[Izacard et~al.(2022)Izacard, Caron, Hosseini, Riedel, Bojanowski,
  Joulin, and Grave]{izacard2022contriever}
Izacard, G., Caron, M., Hosseini, L., Riedel, S., Bojanowski, P., Joulin, A.,
  and Grave, E.
\newblock Unsupervised dense information retrieval with contrastive learning,
  2022.
\newblock URL \url{https://arxiv.org/abs/2112.09118}.

\bibitem[Karpukhin et~al.(2020)Karpukhin, Oguz, Min, Lewis, Wu, Edunov, Chen,
  and Yih]{karpukhin2020dpr}
Karpukhin, V., Oguz, B., Min, S., Lewis, P., Wu, L., Edunov, S., Chen, D., and
  Yih, W.-t.
\newblock Dense passage retrieval for open-domain question answering.
\newblock In Webber, B., Cohn, T., He, Y., and Liu, Y. (eds.),
  \emph{Proceedings of the 2020 Conference on Empirical Methods in Natural
  Language Processing (EMNLP)}, pp.\  6769--6781, Online, November 2020.
  Association for Computational Linguistics.
\newblock \doi{10.18653/v1/2020.emnlp-main.550}.
\newblock URL \url{https://aclanthology.org/2020.emnlp-main.550/}.

\bibitem[Khattab \& Zaharia(2020)Khattab and Zaharia]{khattab2020colbert}
Khattab, O. and Zaharia, M.
\newblock Colbert: Efficient and effective passage search via contextualized
  late interaction over bert.
\newblock In \emph{Proceedings of the 43rd International ACM SIGIR Conference
  on Research and Development in Information Retrieval}, SIGIR '20, pp.\
  39--48, New York, NY, USA, 2020. Association for Computing Machinery.
\newblock ISBN 9781450380164.
\newblock \doi{10.1145/3397271.3401075}.
\newblock URL \url{https://doi.org/10.1145/3397271.3401075}.

\bibitem[Lewis et~al.(2020)Lewis, Perez, Piktus, Petroni, Karpukhin, Goyal,
  K{\"u}ttler, Lewis, Yih, Rockt{\"a}schel, Riedel, and Kiela]{lewis2020rag}
Lewis, P., Perez, E., Piktus, A., Petroni, F., Karpukhin, V., Goyal, N.,
  K{\"u}ttler, H., Lewis, M., Yih, W.-t., Rockt{\"a}schel, T., Riedel, S., and
  Kiela, D.
\newblock Retrieval-augmented generation for knowledge-intensive nlp tasks.
\newblock In \emph{Proceedings of the 34th International Conference on Neural
  Information Processing Systems}, NIPS '20, Red Hook, NY, USA, 2020. Curran
  Associates Inc.
\newblock ISBN 9781713829546.

\bibitem[Li et~al.(2026)Li, Zhang, Wei, Lu, Nie, Lu, Bai, Feng, Zhu, Zhong,
  Zhang, Xie, Choi, Zou, Han, Chen, Lin, Jiang, and Zhang]{dci2026interface}
Li, Z., Zhang, H., Wei, C., Lu, P., Nie, P., Lu, Y., Bai, Y., Feng, S., Zhu,
  H., Zhong, M., Zhang, Y., Xie, J., Choi, Y., Zou, J., Han, J., Chen, W., Lin,
  J., Jiang, D., and Zhang, Y.
\newblock Beyond semantic similarity: Rethinking retrieval for agentic search
  via direct corpus interaction, 2026.
\newblock URL \url{https://arxiv.org/abs/2605.05242}.

\bibitem[Liu et~al.(2024)Liu, Lin, Hewitt, Paranjape, Bevilacqua, Petroni, and
  Liang]{liu2024lost}
Liu, N.~F., Lin, K., Hewitt, J., Paranjape, A., Bevilacqua, M., Petroni, F.,
  and Liang, P.
\newblock Lost in the middle: How language models use long contexts.
\newblock \emph{Transactions of the Association for Computational Linguistics},
  12:\penalty0 157--173, 2024.
\newblock \doi{10.1162/tacl_a_00638}.
\newblock URL \url{https://aclanthology.org/2024.tacl-1.9/}.

\bibitem[Robertson et~al.(1995)Robertson, Walker, Jones, Hancock-Beaulieu, and
  Gatford]{robertson1995bm25}
Robertson, S., Walker, S., Jones, S., Hancock-Beaulieu, M.~M., and Gatford, M.
\newblock Okapi at trec-3.
\newblock In \emph{Overview of the Third Text REtrieval Conference (TREC-3)},
  pp.\  109--126. Gaithersburg, MD: NIST, January 1995.
\newblock URL
  \url{https://www.microsoft.com/en-us/research/publication/okapi-at-trec-3/}.

\bibitem[Schick et~al.(2023)Schick, Dwivedi-Yu, Dess{\'i}, Raileanu, Lomeli,
  Hambro, Zettlemoyer, Cancedda, and Scialom]{schick2023toolformer}
Schick, T., Dwivedi-Yu, J., Dess{\'i}, R., Raileanu, R., Lomeli, M., Hambro,
  E., Zettlemoyer, L., Cancedda, N., and Scialom, T.
\newblock Toolformer: language models can teach themselves to use tools.
\newblock In \emph{Proceedings of the 37th International Conference on Neural
  Information Processing Systems}, NIPS '23, Red Hook, NY, USA, 2023. Curran
  Associates Inc.

\bibitem[Sen et~al.(2026{\natexlab{a}})Sen, Kasturi, Lumer, Gulati, and
  Subbiah]{sen2026grep}
Sen, S., Kasturi, A., Lumer, E., Gulati, A., and Subbiah, V.~K.
\newblock Is grep all you need? how agent harnesses reshape agentic search,
  2026{\natexlab{a}}.
\newblock URL \url{https://arxiv.org/abs/2605.15184}.

\bibitem[Sen et~al.(2026{\natexlab{b}})Sen, Lumer, Gulati, and
  Subbiah]{lumer2026chronos}
Sen, S., Lumer, E., Gulati, A., and Subbiah, V.~K.
\newblock Chronos: Temporal-aware conversational agents with structured event
  retrieval for long-term memory, 2026{\natexlab{b}}.
\newblock URL \url{https://arxiv.org/abs/2603.16862}.

\bibitem[Thakur et~al.(2021)Thakur, Reimers, Rücklé, Srivastava, and
  Gurevych]{thakur2021beir}
Thakur, N., Reimers, N., Rücklé, A., Srivastava, A., and Gurevych, I.
\newblock Beir: A heterogenous benchmark for zero-shot evaluation of
  information retrieval models, 2021.
\newblock URL \url{https://arxiv.org/abs/2104.08663}.

\bibitem[Trivedi et~al.(2023)Trivedi, Balasubramanian, Khot, and
  Sabharwal]{trivedi2023ircot}
Trivedi, H., Balasubramanian, N., Khot, T., and Sabharwal, A.
\newblock Interleaving retrieval with chain-of-thought reasoning for
  knowledge-intensive multi-step questions, 2023.
\newblock URL \url{https://arxiv.org/abs/2212.10509}.

\bibitem[Wang et~al.(2024)Wang, Wang, Gao, Zhang, Wu, Xu, Shi, Wang, Li, Qian,
  Yin, Lv, Zheng, and Huang]{wang2024ragbestpractices}
Wang, X., Wang, Z., Gao, X., Zhang, F., Wu, Y., Xu, Z., Shi, T., Wang, Z., Li,
  S., Qian, Q., Yin, R., Lv, C., Zheng, X., and Huang, X.
\newblock Searching for best practices in retrieval-augmented generation.
\newblock In Al-Onaizan, Y., Bansal, M., and Chen, Y.-N. (eds.),
  \emph{Proceedings of the 2024 Conference on Empirical Methods in Natural
  Language Processing}, pp.\  17716--17736, Miami, Florida, USA, November 2024.
  Association for Computational Linguistics.
\newblock \doi{10.18653/v1/2024.emnlp-main.981}.
\newblock URL \url{https://aclanthology.org/2024.emnlp-main.981/}.

\bibitem[Xia et~al.(2024)Xia, Deng, Dunn, and Zhang]{xia2024agentless}
Xia, C.~S., Deng, Y., Dunn, S., and Zhang, L.
\newblock Agentless: Demystifying llm-based software engineering agents, 2024.
\newblock URL \url{https://arxiv.org/abs/2407.01489}.

\bibitem[Yang et~al.(2024)Yang, Jimenez, Wettig, Lieret, Yao, Narasimhan, and
  Press]{yang2024sweagent}
Yang, J., Jimenez, C., Wettig, A., Lieret, K., Yao, S., Narasimhan, K., and
  Press, O.
\newblock Swe-agent: Agent-computer interfaces enable automated software
  engineering.
\newblock \emph{Advances in Neural Information Processing Systems}, 37, 2024.
\newblock ISSN 1049-5258.

\bibitem[Yao et~al.(2023)Yao, Zhao, Yu, Du, Shafran, Narasimhan, and
  Cao]{yao2023react}
Yao, S., Zhao, J., Yu, D., Du, N., Shafran, I., Narasimhan, K., and Cao, Y.
\newblock React: Synergizing reasoning and acting in language models, 2023.
\newblock URL \url{https://arxiv.org/abs/2210.03629}.

\bibitem[Yu(2025)]{yu2025grepvsgraph}
Yu, F.~J.
\newblock Grep vs graph: Agentic search is powerful, but enterprise {AI} needs
  governed knowledge.
\newblock
  \url{https://medium.com/@yu-joshua/grep-vs-graph-agentic-search-is-powerful-but-enterprise-ai-needs-governed-knowledge-8de709c31451},
  2025.
\newblock Accessed 2026-08-03.

\end{thebibliography}
\bibliographystyle{icml2026}

\newpage
\appendix
\onecolumn

\section{Converter Comparison}
\label{app:conversion}

Section~\ref{sec:conversion} argues that PDF$\rightarrow$Markdown conversion sets a
ceiling on what any retriever can recover. That ceiling is a property of the
converter, so the converter was chosen before any retrieval system was built,
and the choice is documented here rather than asserted.

Four backends were run over the full 780-page source. Three completed and
reached the back cover; \texttt{docling} never finished (below).

\begin{table}[h]
\centering
\caption{Converter comparison on the full document. The rating is our
judgement; the counts beneath it are what drove it.}
\label{tab:converters}
\begin{tabular}{llp{8.4cm}}
\toprule
backend & verdict & failure mode \\
\midrule
\texttt{pymupdf4llm} & \textbf{chosen} & Keeps table columns aligned. Known
defects accepted: a mangled table of contents and a leading empty column on
every row. Neither touches the financial statements. \\
\texttt{cloudconvert} & rejected & Collapses whole tables into a single
\texttt{<br>}-joined cell --- up to 30 printed rows in one cell. Salvageable in
principle, not usable as given. \\
\texttt{markitdown}  & rejected & Severs row labels from their figures. The loss
is unrecoverable: no downstream process can reattach a label to the right row. \\
\texttt{docling}     & unrated & Never completed a run. \\
\bottomrule
\end{tabular}
\end{table}

\paragraph{The deciding evidence.} The same statement row, as rendered by each
backend:

\begin{quote}\small\ttfamily
\begin{tabular}{@{}ll@{}}
pymupdf4llm  & \verb!||Co-operation|26.23|15.93|! \\
             & \emph{columns intact} \\[2pt]
cloudconvert & \verb!|**(B)**<br>Housing<br>74.48<br>18.48<br>...|! \\
             & \emph{30 rows in one cell} \\[2pt]
markitdown   & \verb!Co-operation! \\
             & \verb!|  |  |  | 26.23 | 15.93 |! \\
             & \emph{label orphaned from its figures} \\
\end{tabular}
\end{quote}

\paragraph{Supporting counts.} \texttt{cloudconvert} emits 83{,}535
\texttt{<br>} tokens and 2{,}459 collapsed mega-cells against
\texttt{pymupdf4llm}'s 18{,}319 and 842; 1{,}506 of its rows are only two
columns wide, which is the collapse signature. \texttt{markitdown} fragments
5{,}547 data rows across 2{,}589 tables, about 2.1 rows per table.
\texttt{pymupdf4llm}'s column counts peak at 5--10, matching the real
statements.

\paragraph{Two measures that look decisive and are not.} Digit counts mislead:
\texttt{cloudconvert} carries roughly 39{,}000 more digit characters, but all
three backends recover the same $\approx$15{,}940 \emph{distinct} numbers --- the
surplus is repetition introduced by cell collapse, not extra information.
Encoding fidelity is a wash: 2 bad glyphs in \texttt{pymupdf4llm}, 2 in
\texttt{markitdown}, 0 in \texttt{cloudconvert}, across about 2\,MB. Had we
scored converters on either metric we would have chosen \texttt{cloudconvert},
whose output is the least usable of the three that completed.

\paragraph{On \texttt{docling}.} Every run died with \texttt{std::bad\_alloc}
during preprocessing. The cause is its default v1 parse backend compounded by
Windows commit-limit exhaustion: at the point of failure the machine had
15.8\,GB of free RAM but only 0.4\,GB of commit headroom. The failing page
drifts with available memory --- page 4 on one run, page 7 on another --- so
this is an environment interaction, not a poisoned page, and we record it as
unrated rather than as a quality judgement.

\paragraph{Caveat.} These ratings are ours, applied to one document, by readers
who knew which backend they had already been using. They are reported so the
conversion ceiling in Section~\ref{sec:conversion} can be attributed to a specific
artifact, not offered as a converter benchmark.

\section{Implementation Details}
\label{app:details}

\subsection{The four operations}

\READ{} exposes four tools over the Model Context Protocol. The server is
read-only and jailed to a root directory: every path is resolved and rejected if
it escapes that root, so no trajectory can read outside the corpus.

\begin{description}[leftmargin=*,topsep=2pt,itemsep=3pt]
\item[\texttt{memory\_list}] Enumerate files under an optional subdirectory.
\item[\texttt{memory\_outline}] Return the heading structure of a file, each
heading with its line number.
\item[\texttt{memory\_grep}] Search line by line. Takes \texttt{query} and
optional \texttt{path}, \texttt{context} (0--10 lines, default 2), and
\texttt{max\_matches} (default 25). Returns \texttt{line\_no:} prefixed matches
with surrounding context.
\item[\texttt{memory\_read}] Return lines \texttt{start\_line} to
\texttt{end\_line} of a file, 1-indexed and inclusive.
\end{description}

The design point is that \texttt{memory\_grep} returns \emph{line numbers} and
\texttt{memory\_read} accepts them, so the agent composes locate-then-read
without any intermediate representation the user cannot inspect. Line numbers
are 1-indexed everywhere, matching the released MCP server, so a trajectory can
be replayed against the document by hand.

\subsection{Normalized matching}

A query is matched against two shadow forms of every line, and succeeds if
either contains it. Both start from a base form: Unicode NFKC normalization,
emphasis markers (\texttt{*}, \texttt{\_}) stripped, case folded.

\begin{description}[leftmargin=*,topsep=2pt,itemsep=2pt]
\item[spaced form] Table-cell boundaries (\texttt{|}) replaced by spaces and
whitespace collapsed. This makes cell structure invisible to matching while
keeping word boundaries.
\item[squashed form] All cell boundaries, whitespace, and thousands separators
removed. This is what lets \texttt{"Fiscal Deficit"} match the converted
\verb!|**Fiscal**| **Defi**|**cit**|!, where the converter split the word across
two cells, and \texttt{"48964.89"} match the printed \texttt{48,964.89}.
\end{description}

An empty query after normalization is an error rather than a match-everything.
\Cref{sec:normalization} measures what each form contributes.

\subsection{The table-aware chunker}

The chunker is our steelman for the dense and hybrid baselines, not part of
\READ. It packs lines up to \texttt{chunk\_chars}, and at every boundary
prepends a context header recovered by scanning \emph{backwards} from the
boundary for two things: the nearest unit declaration
(\texttt{(\rupee{} in lakh)} and variants) and the nearest fiscal-year or
as-at-date header. Both are emitted in document order so the carried header
reads as it does in the book.

Two details matter. First, the header search looks backwards rather than
carrying the table run's first rows: in this corpus the unit declaration
usually sits \emph{above} the table run, so carrying the run's own first rows
propagates decoration and not meaning. Fixing this moved the fraction of
unitless numeric chunks from 18.0\% to 0.3\% and is the difference between the
steelman working and appearing to work. Second, the seed budget is capped at
$\max(200, \texttt{chunk\_chars}/2)$ and each carried line at
$\max(120, \texttt{chunk\_chars}/4)$, so a wide header cannot crowd out the
chunk body it is meant to explain.

The chunker is forward-only. An earlier rewind-based implementation of overlap
could fail to make forward progress on long table runs and exhaust memory; the
current one always advances. The chunker version participates in the index cache
key, so a change to the algorithm cannot be silently served from a stale index.

\subsection{Models and decoding}

All systems use \texttt{gemini-2.5-pro} as the generator through the
\texttt{google-genai} SDK, with a manual tool loop rather than the SDK's
automatic one so that every call, its arguments, and the exact text it exposed
are logged. Dense retrieval uses \texttt{BAAI/bge-base-en-v1.5}; hybrid adds
\texttt{cross-encoder/ms-marco-MiniLM-L-6-v2} as a reranker over the fused
candidate set.

One decoding detail is not cosmetic. Gemini charges thinking tokens against
\texttt{max\_output\_tokens}. Under a dynamic thinking budget the model can
spend the entire allowance reasoning and emit zero output tokens, returning an
empty answer with \texttt{stop\_reason=MAX\_TOKENS} --- which a grader scores as
a wrong answer rather than as a harness failure. This affected 2 of 175 cells in
an early run before we found it. Thinking is now capped at half the output
budget. We report it because the failure is silent, is invisible in accuracy
numbers, and would bias whichever system is given the least generous budget.

\section{Benchmark Details}
\label{app:data}

\subsection{Composition}

The benchmark holds 51 questions with verified gold answers, drawn from six
categories (\Cref{tab:bench-comp}). Gold evidence is recorded as line numbers
into the converted Markdown, from which printed page numbers are derived.

\begin{table}[h]
\centering
\caption{Benchmark composition. ``Derived'' questions have answers that appear
nowhere in the document and must be computed.}
\label{tab:bench-comp}
\begin{tabular}{lrp{7.6cm}}
\toprule
category & $n$ & what it tests \\
\midrule
lookup            & 24 & Retrieve one printed figure from a named row and column. \\
arithmetic        &  9 & Compute a value that is not printed, from operands that are. \\
navigation        &  6 & Answer about document structure: which statement, which page, which unit. \\
aggregation       &  5 & Combine two or more printed figures into one answer. \\
unanswerable      &  5 & Decline. The document does not contain the answer. \\
conversion\_limited & 2 & The evidence exists in the PDF but was damaged in conversion (\Cref{sec:conversion}). \\
\midrule
total             & 51 & \\
\bottomrule
\end{tabular}
\end{table}

By difficulty the split is 32 hard, 13 medium, 6 easy. Ten questions carry
\emph{derived} answers, validated as described below. Gold evidence spans
printed pages 2 to 544 with a median of 17; 14 questions have all their evidence
past printed page 100, and the widest single question spans 3{,}202 lines of
Markdown. Fifteen questions sit on pages printed in crore and fifteen on pages
printed in lakh, which is what makes the unit-inheritance failure in
\Cref{sec:corpus} testable rather than hypothetical.

\subsection{Annotation protocol}

Questions were drafted in batches against the converted document, and every
batch was validated by a script before being admitted. Three checks run on each
batch, and a batch that fails any of them is not written:

\begin{enumerate}[leftmargin=*,topsep=2pt,itemsep=2pt]
\item \textbf{Presence.} Every expected figure, and every operand of a derived
answer, must appear within one line of a declared gold line. This catches a
figure transcribed from the PDF that the converter rendered differently.

\item \textbf{Column mapping by arithmetic.} The hazard in a wide financial
table is not misreading a digit but reading the right digit from the wrong
column --- a mistake that survives proofreading, because the drafter re-reads
the table the same wrong way. Where the table carries an internal identity, we
check it on every source row rather than trusting the header. In Statement 4B
that identity is Revenue $+$ Capital $=$ Total; in Statement 17 it is
opening balance $+$ additions $-$ discharges $=$ closing balance; Statement 18
prints its own formula, $(3+4)-(5+6)=7$. All fifteen source rows used by the
deep-document batch reconcile.

\item \textbf{Derived-answer validation.} An arithmetic question is only
arithmetic if its answer is not already printed somewhere. The validator
searches the whole document for every printed form of the expected answer ---
including Indian digit grouping, so $4{,}72{,}397.46$ and $472{,}397.46$ are
both checked --- and rejects the question if it finds one. This caught a drafted
question whose answer, 40.93, turned out to be printed elsewhere in the volume:
it was a lookup wearing an arithmetic costume, and was replaced.
\end{enumerate}

Thirteen further questions have been drafted but not yet cleared these checks;
they are excluded from all reported results, and the runner refuses to score
them unless explicitly overridden.

\subsection{A note on scale and on who annotated}

Fifty-one questions is small. At this size a paired test resolves only large
differences, and we report the minimum detectable effect alongside every
comparison rather than treating a null result as evidence of equivalence
(\Cref{sec:metrics}). The questions were drafted by the authors, who also built
the system under test --- the standard concern applies, that a benchmark built
by a system's authors tends to reward that system. Two things partly offset it:
the gold answers are checked mechanically against the document rather than by
the drafter's own reading, and the categories where \READ{} is weakest are
included and reported rather than dropped. It remains a limitation, and is
stated as one in \Cref{sec:limitations}.

\subsection{Licensing}

The source document, \emph{Finance Accounts 2024--25, Government of Gujarat,
Volume I}, is a public record published by the Comptroller and Auditor General
of India. The benchmark file distributes questions, gold answers, and line
references --- not the document text --- and the conversion pipeline needed to
reproduce the Markdown is released with the code.

\section{Dense-RAG Sweep}
\label{app:sweep}

A comparison against a badly configured baseline proves nothing. This appendix
asks whether the dense configuration reported in \Cref{sec:results} ---
2{,}000-character table-aware chunks, $k{=}8$ --- is the best available to dense
retrieval on this document, or whether we hobbled it. We swept chunk size and
retrieval depth over the same 51 verified questions, the same backbone, and the
same graders: 6 configurations $\times$ 51 questions $=$ 306 further cells, all
of which completed without error.

\begin{table}[t]
\centering
\small
\begin{tabular}{lccc}
\toprule
& \multicolumn{3}{c}{top-$k$} \\
\cmidrule(lr){2-4}
Chunk size & 4 & 8 & 16 \\
\midrule
2{,}000 & 17.6 & 17.6$^\dagger$ & \textbf{35.3} \\
4{,}000 & 19.6 & 21.6 & 29.4 \\
\bottomrule
\end{tabular}
\caption{Dense accuracy (\%) over chunk size and retrieval depth, $n{=}51$ per
cell. $^\dagger$This is the configuration reported as \emph{dense} in
\Cref{tab:main}, where it scored 15.7\%; the one-question difference is
run-to-run nondeterminism in the backbone, not a change of setting.}
\label{tab:sweep}
\end{table}

\begin{figure}[t]
\centering
\includegraphics[width=0.85\linewidth]{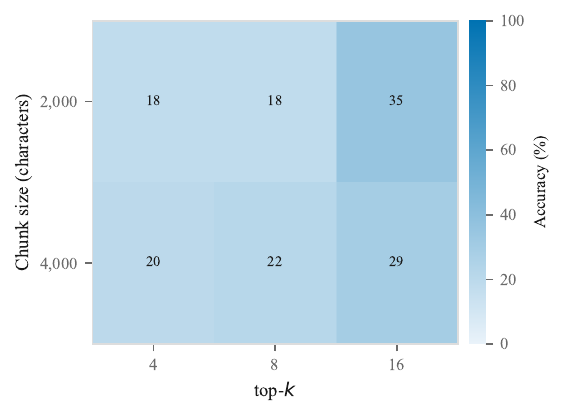}
\caption{The same grid. Colour encodes accuracy; every cell is annotated, so the
figure is readable without the ramp.}
\label{fig:chunk_sweep}
\end{figure}

Three things follow.

\textbf{Depth is the lever; chunk size is not.} Going from $k{=}4$ to $k{=}8$
moves accuracy by at most two points. Going from $k{=}8$ to $k{=}16$ roughly
doubles it at 2{,}000 characters (17.6\% to 35.3\%). Chunk size, over the
two-fold range we ran, moves accuracy by less than four points at fixed $k$ and
does not even do so monotonically --- 4{,}000 is better at $k{=}4$ and $k{=}8$,
worse at $k{=}16$. This is consistent with \Cref{sec:corpus}: the failure is
that a retrieved chunk does not carry the header governing its figures, and the
only reliable remedy available to a top-$k$ system is to retrieve more chunks
and hope one of them does.

\textbf{The reported baseline is not dense's ceiling, and we say so.} The best
configuration we measured is 2{,}000 characters at $k{=}16$, scoring 35.3\% ---
more than double the 15.7\% carried through \Cref{sec:results}. Any reading of
this paper should use 35.3\% as dense retrieval's honest best on this benchmark.
We did not restate \Cref{tab:main} around it because $k{=}8$ is the value used
by every other retrieval baseline in the paper, and changing it for dense alone
would break the matched comparison; the effect of that choice is to understate
dense, and it is corrected here.

\textbf{\READ{} still separates from every configuration.} Paired exact McNemar
against \READ's 58.8\%, Holm-corrected across the six comparisons, rejects in
all six cases. Against the best configuration the margin narrows from 43.1 to
23.5 points and remains significant.

\begin{table}[t]
\centering
\small
\begin{tabular}{lrcr}
\toprule
Configuration & $\Delta$ (pp) & 95\% CI & $p_{\text{Holm}}$ \\
\midrule
2{,}000, $k{=}16$ & $+23.5$ & [\phantom{0}5.9,\,39.2] & 0.017 \\
4{,}000, $k{=}16$ & $+29.4$ & [11.8,\,47.1] & 0.012 \\
4{,}000, $k{=}8$  & $+37.3$ & [19.6,\,53.0] & 0.0009 \\
4{,}000, $k{=}4$  & $+39.2$ & [23.5,\,54.9] & 0.0005 \\
2{,}000, $k{=}4$  & $+41.2$ & [23.5,\,58.8] & 0.0005 \\
2{,}000, $k{=}8$  & $+41.2$ & [23.5,\,56.9] & 0.0003 \\
\bottomrule
\end{tabular}
\caption{\READ{} against every swept dense configuration. Positive $\Delta$
favours \READ. All six reject after Holm correction.}
\label{tab:sweep_paired}
\end{table}

Note that the 23.5-point margin is smaller than the 27.7-point minimum
detectable difference quoted in \Cref{sec:results}, yet reaches $p=0.017$. This
is not a contradiction: that figure is an \emph{a priori}, unpaired
approximation at a base rate of one half, whereas the test actually performed is
paired and conditions on discordant questions alone. We report it because it
sets expectations for what a benchmark of this size can see, not as a criterion
for declaring a gap unresolvable.

Finally, depth is not free. At 2{,}000 characters, $k{=}16$ costs
\$0.029 per question against \$0.021 at $k{=}8$, a 41\% increase, for a system
that still trails \READ{} by 23.5 points at 51\% of \READ's cost. The whole
sweep cost \$8.07.

\textbf{Limitation.} We did not run chunk sizes below 2{,}000. On a CPU-only
machine each new chunk size requires re-embedding the corpus, and the
800-character index --- 4{,}083 chunks against 534 at 4{,}000 --- was estimated
at over an hour and was twice killed by the host before completing.
\Cref{tab:chunking} predicts that setting is strictly worse rather than better:
at 800 characters, 67.4\% of fixed-size chunks carry no unit declaration, the
highest of any size we measured. We therefore expect it to lie below the grid
reported here, but we did not measure it, and a reader who wants that cell
should treat it as open.

\section{Example Trajectories}
\label{app:trajectories}
Both systems below were asked the question from \Cref{sec:intro}, on the same
document, with the same backbone:

\begin{quote}\itshape
Based on the Statement of Receipts and Disbursements, what is the mathematical
difference between the Fiscal Deficit and the Revenue Surplus for the year
2024--25?
\end{quote}

The two operands sit 47 lines apart in Statement No.~2, and their difference
(\rupee{}30{,}021.95 crore) is printed nowhere in the volume.

\subsection{Dense retrieval: one shot, and it declines}

\begin{quote}\small\ttfamily
\textbf{dense\_retrieve(k=8)} $\rightarrow$ 8 chunks \\[4pt]
\textbf{ANSWER:} The provided excerpts state the Revenue Surplus for 2024-25 but
do not contain a figure for the Fiscal Deficit for that year. Therefore, the
mathematical difference between the two cannot be calculated from the
information supplied. The Revenue Surplus for 2024-25 is \rupee{}18,942.94
crore. \\[4pt]
\textbf{EVIDENCE:} Chunk 90, line 2002 \dots
\end{quote}

\emph{11\,s, \$0.021, one retrieval call, scored incorrect.} The retriever
surfaced one operand and not the other, and the generator behaved correctly
given what it was handed: it declined. Nothing in the trajectory is a
reasoning failure. The partition was fixed before the query, both operands were
never going to arrive together, and the pipeline has no way to ask for more.

\subsection{\READ: eight operations, then the answer}

\begin{quote}\small\ttfamily
memory\_grep("Statement of Receipts and Disbursements") $\rightarrow$ 5 hits \\
memory\_grep("Fiscal Deficit") $\rightarrow$ 2 hits \\
memory\_outline() $\rightarrow$ 190 headings \\
memory\_grep("Fiscal Deficit") $\rightarrow$ 2 hits \\
memory\_grep("Revenue Surplus") $\rightarrow$ 5 hits \\
memory\_grep("STATEMENT No. 2") $\rightarrow$ 17 hits \\
memory\_read(lines 365--450) $\rightarrow$ 86 lines \\
memory\_read(lines 1023--1120) $\rightarrow$ 98 lines \\[4pt]
\textbf{ANSWER:} \dots the Fiscal Deficit for the year 2024-25 is
\rupee{}48,964.89 crore and the Revenue Surplus for the same year is
\rupee{}18,942.94 crore. The mathematical difference between these two figures
is \rupee{}30,021.95 crore.
\end{quote}

\emph{49\,s, \$0.097, eight operations, scored correct.} \READ{} costs
4.6$\times$ more and takes 4.5$\times$ longer, and this is the honest shape of
the trade.

What the trajectory shows is the argument of the paper made concrete. The agent
locates the statement by name, fails to get enough from two narrow greps, falls
back to the outline to orient itself, returns to the greps now knowing where it
is, and finally reads two explicit line spans --- widening its own context twice
in response to what the document told it. Each step is a pure function of the
document text: any reader can re-execute
\texttt{memory\_read(365--450)} and see the same 86 lines. The final answer
cites line numbers that a reviewer can check by hand, which is the property a
similarity score cannot offer.

Note also the redundant second \texttt{memory\_grep("Fiscal Deficit")}, issued
after the outline call returned nothing useful. \READ{} trajectories contain
wasted operations. They are visible, which is the point: the failure modes of
this system are legible in a way that a top-$k$ ranking is not.

\end{document}